\documentclass[letterpaper]{article} % DO NOT CHANGE THIS

\usepackage{aaai2026}
\usepackage{times}  % DO NOT CHANGE THIS
\usepackage{helvet}  % DO NOT CHANGE THIS
\usepackage{courier}  % DO NOT CHANGE THIS
\usepackage[hyphens]{url}  % DO NOT CHANGE THIS
\usepackage{graphicx} % DO NOT CHANGE THIS
\usepackage{natbib}  % DO NOT CHANGE THIS AND DO NOT ADD ANY OPTIONS TO IT
\usepackage{caption} % DO NOT CHANGE THIS AND DO NOT ADD ANY OPTIONS TO IT
\usepackage{amsmath}
\DeclareMathOperator*{\argmax}{arg\,max}
\usepackage{algorithm}
\usepackage{algorithmic}
\usepackage{xcolor}
\usepackage{colortbl}
\usepackage{xspace}
\usepackage{amssymb}
\usepackage{multirow}
\usepackage{booktabs}
\usepackage{array}
\usepackage{tabularx}
\usepackage{fancyhdr}
\definecolor{LightCyan}{rgb}{0.88,1,1}
\definecolor{LightRed}{rgb}{1,0.5,0.5}
\definecolor{LightYellow}{rgb}{1,1,0.88}
\definecolor{Grey}{rgb}{0.75,0.75,0.75}
\definecolor{DarkGrey}{rgb}{0.55,0.55,0.55}
\definecolor{LightGreen}{rgb}{0.0, 0.70, 0.0}

\nocopyright
\usepackage{newfloat}
\usepackage{listings}
\DeclareCaptionStyle{ruled}{labelfont=normalfont,labelsep=colon,strut=off} % DO NOT CHANGE THIS
\floatstyle{ruled}
\newfloat{listing}{tb}{lst}{}
\floatname{listing}{Listing}
\usepackage{hyperref} 
\usepackage{url}

\title{Thinking Inside the Mask: In-Place Prompting in Diffusion LLMs}
\author {
    Xiangqi Jin\textsuperscript{\rm 1}, 
    Yuxuan Wang\textsuperscript{\rm 2}, 
    Yifeng Gao\textsuperscript{\rm 1}, 
    Zichen Wen\textsuperscript{\rm 1,\rm 3}, \\
    Biqing Qi\textsuperscript{\rm 3}, 
    Dongrui Liu\textsuperscript{\rm 3}, 
    Linfeng Zhang\textsuperscript{\rm 1}\thanks{Corresponding author (zhanglinfeng@sjtu.edu.cn).}
}
\affiliations {
    \textsuperscript{\rm 1}School of Artificial Intelligence, Shanghai Jiao Tong University\\
    \textsuperscript{\rm 2}Xidian University\\
    \textsuperscript{\rm 3}Shanghai AI Laboratory 
}

\usepackage{bibentry}
\begin{document}

\maketitle

\begin{abstract}
Despite large language models (LLMs) have achieved remarkable success, their \textit{prefix-only} prompting paradigm and sequential generation process offer limited flexibility for bidirectional information. Diffusion large language models (dLLMs) present new opportunities through their bidirectional attention mechanisms and iterative refinement processes, enabling more flexible \textit{in-place} prompting strategies. We introduce ICE (\underline{\textbf{I}}n-Place \underline{\textbf{C}}hain-of-Thought Prompting with \underline{\textbf{E}}arly Exit), a novel framework that transforms prefix-only prompting into in-place prompting specifically designed for dLLMs. ICE integrates in-place prompts directly within masked token positions during iterative refinement and employs a confidence-aware early exit mechanism to significantly reduce computational overhead. Extensive experiments demonstrate ICE's effectiveness, achieving up to 17.29\% accuracy improvement with 4.12$\times$ speedup on GSM8K, and up to 276.67$\times$ acceleration on MMLU while maintaining competitive performance. Our code is available at \url{https://github.com/Lueci4er/ICE}.
\end{abstract}


\section{Introduction}\label{sec:introduction}

Large language models (LLMs)~\citep{zhao2023survey} have revolutionized natural language processing, with autoregressive (AR) models dominating the landscape through their sequential, left-to-right token generation paradigm~\cite{brown2020language,touvron2023llama}. AR models are fundamentally constrained by \textit{prefix-only} prompting and sequential generation. Diffusion large language models (dLLMs)~\citep{dream2025,nie2025large,Zhu2025LLaDA1V,yang2025mmada} offer non-autoregressive alternatives through iterative masked token refinement~\cite{austin2021structured,lou2023discrete}. Crucially, dLLMs present new opportunities through their bidirectional attention mechanisms and iterative refinement processes, enabling more flexible \textit{in-place} prompting strategies that can embed information directly within masked token positions~\citep{wen2025devil}. LLaDA matches GPT-3.5 performance~\cite{nie2025large} and Mercury achieves 1000+ tokens/second~\cite{liu2025mercury}, yet reasoning capabilities in dLLMs remain underexplored.

\begin{figure}[t!]
    \centering
    \includegraphics[width=0.9\columnwidth]{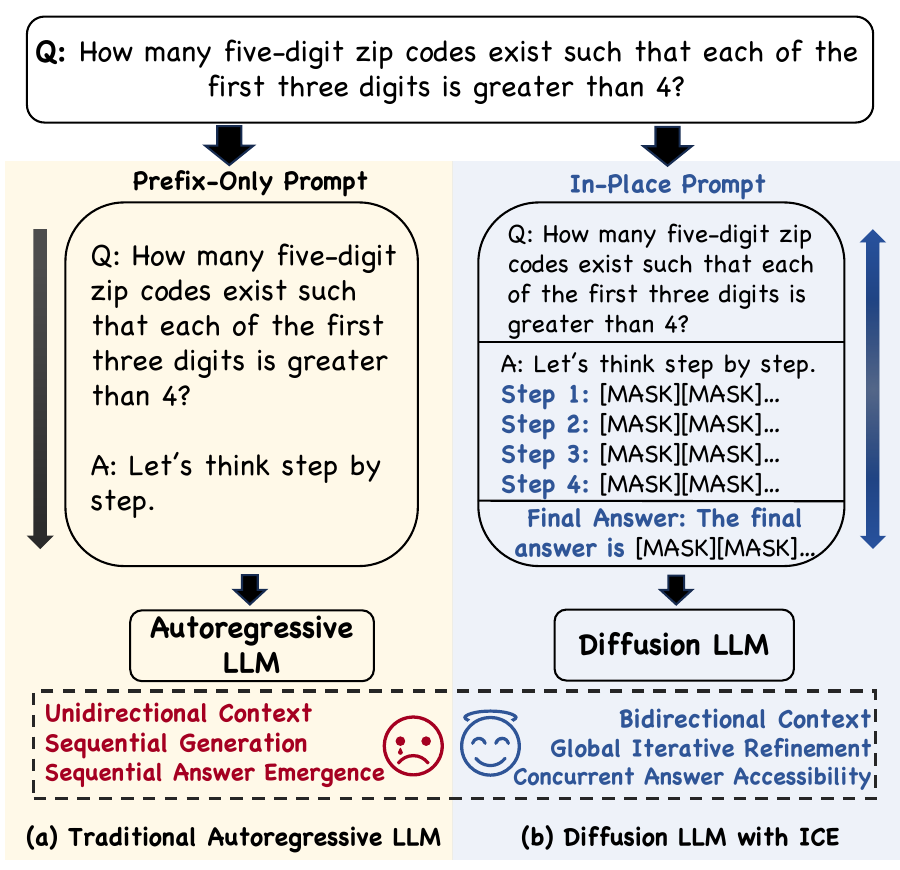} 
    \caption{Prompt construction in \textbf{(a) Autoregressive LLMs} vs. \textbf{(b) Diffusion LLMs}. Autoregressive LLMs employ unidirectional context with prefix-only prompting, while dLLMs leverage bidirectional context modeling, enabling in-place prompting and concurrent answer accessibility.}
    \label{fig:teaser}
\end{figure}

While Chain-of-Thought (CoT) prompting has proven highly effective for AR models by decomposing complex problems into intermediate reasoning steps~\cite{wei2022chainofthought}, the bidirectional and iterative nature of dLLMs enables fundamentally different approaches. Unlike AR models that treat reasoning as sequential prefix conditioning, dLLMs can embed reasoning directly within the generation process itself with in-place prompting (Figure~\ref{fig:teaser}). Moreover, AR models exhibit \textit{sequential answer emergence}, where answers remain inaccessible until the completion of sequential generation, while dLLMs enable \textit{concurrent answer accessibility} through their bidirectional context modeling, allowing intermediate visibility of answer content during the iterative refinement process. This architectural distinction creates opportunities for novel confidence-aware optimization strategies that can monitor answer during generation.

\begin{figure*}[t!]
    \centering
    \includegraphics[width=\textwidth]{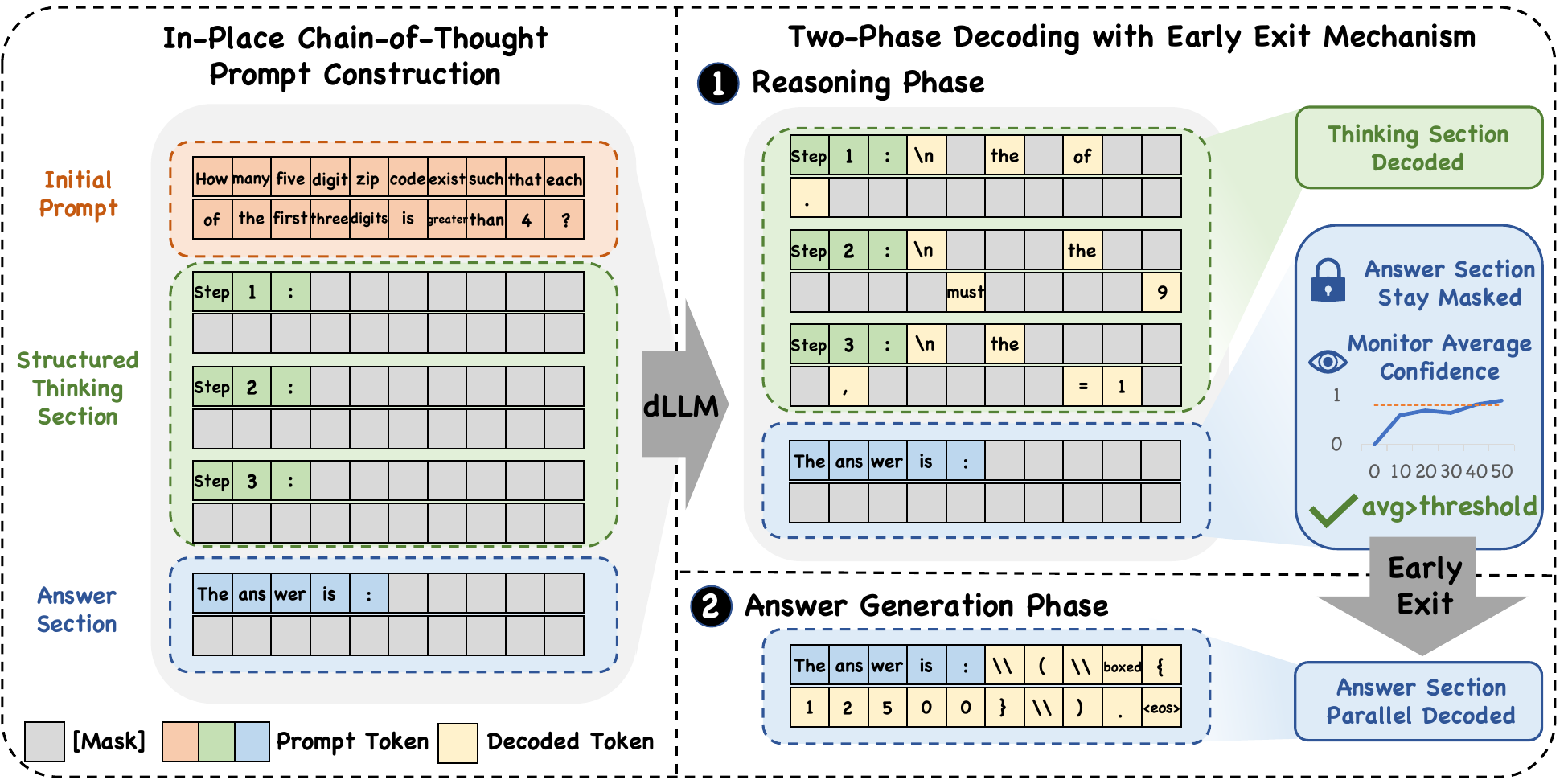} 
    \caption{Overview of the ICE framework. ICE integrates two key components: \textbf{(1) In-Place Chain-of-Thought Prompting}, which embeds structured step-by-step reasoning templates directly into the prompt, and \textbf{(2) Two-Phase Decoding with Confidence-Aware Early Exit Mechanism}. During the \textit{Reasoning Phase}, the model iteratively decodes the thinking section while monitoring the average confidence of the masked answer section. Once the confidence threshold is reached, the framework transitions to the \textit{Answer Generation Phase}, decoding all answer tokens in parallel to produce the final response.}
    \label{fig:main_overview}
\end{figure*}

To address these opportunities, we propose ICE (\underline{\textbf{I}}n-Place \underline{\textbf{C}}hain-of-Thought Prompting with \underline{\textbf{E}}arly Exit), a novel framework that enhances both reasoning capabilities and inference efficiency in dLLMs (Figure~\ref{fig:main_overview}). Our central insight is that the iterative generation process of dLLMs provides a unique opportunity to embed reasoning steps directly within the generation process, transforming reasoning from external preprocessing into an integral component of the generation mechanism. ICE introduces two key innovations:

\noindent \textbf{In-Place Chain-of-Thought Prompting:} This approach integrates reasoning steps directly into masked token positions during iterative refinement. It exploits the bidirectional nature of dLLMs by structuring the generation sequence into distinct thinking and answer sections, with explicit step-by-step reasoning templates embedded within the thinking section. This enables enhanced reasoning performance while preserving parallel generation advantages.

\begin{figure}[t!]
    \centering
    \includegraphics[width=0.9\columnwidth]{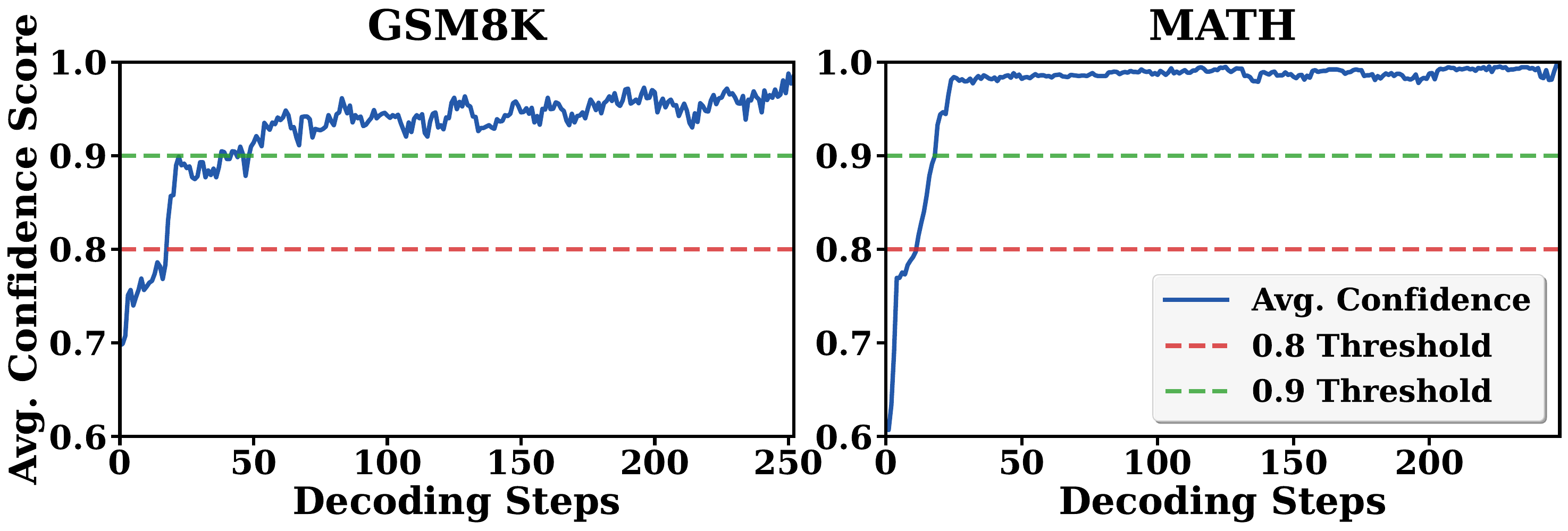} 
    \caption{Average confidence of answer section on GSM8K and MATH during generation. The model's confidence in the answer section rapidly converges to a high level and remains stable throughout subsequent iterations, indicating that the model internally determines the correct answer early in the process while continuing to refine the reasoning trace.}
    \label{fig:confidence_trends}
\end{figure}

\noindent \textbf{Two-Phase Decoding with Early Exit Mechanism:} Motivated by a crucial empirical observation, we design a confidence-aware inference strategy that capitalizes on the distinct refinement patterns between reasoning and answer components. Through systematic analysis of iterative refinement dynamics, we uncover a distinctive behavioral pattern in dLLMs: model confidence in answer tokens converges rapidly to high levels and maintains stability, while the reasoning section continues undergoing refinement (Figure~\ref{fig:confidence_trends}). This observation reveals that models often internally determine correct answers significantly earlier than the completion of explicit reasoning traces. Leveraging this insight, ICE implements a two-phase decoding approach that enables parallel decoding of all answer tokens while effectively reducing redundant computation.

Extensive experimental validation demonstrates ICE's effectiveness across diverse reasoning benchmarks. On mathematical reasoning tasks, ICE achieves up to 17.29\% accuracy improvement with 4.12$\times$ speedup on GSM8K, alongside consistent gains on MATH. For knowledge-intensive tasks, ICE delivers up to 276.67$\times$ acceleration on MMLU and speedups exceeding 40$\times$ on GPQA while maintaining competitive accuracy. Notably, ICE is  compatible with existing acceleration techniques, such as dLLM-Cache~\cite{liu2025dllmcacheacceleratingdiffusionlarge}, enabling cumulative benefits when combined.

Our contributions are threefold:
\begin{itemize}
    \item We introduce the first in-place prompting framework for dLLMs, embedding prompts directly in masked tokens to improve both accuracy and efficiency.
    \item We develop a two-phase decoding strategy with early exit mechanism that significantly reduces inference latency while maintaining generation quality.
    \item We provide comprehensive empirical evidence demonstrating ICE's effectiveness, establishing that architectural alignment between reasoning patterns and generation mechanisms can yield synergistic benefits.
\end{itemize}

\section{Related Work}\label{sec:related_work}
%%%%%% More citation.
\noindent \textbf{Diffusion LLMs.}
Recent diffusion large language models (dLLMs)~\cite{nie2025large, austin2021program, lou2023discrete, shi2024simplified, liu2025longllada}, notably LLaDA, represent a paradigm shift from autoregressive generation by employing bidirectional attention mechanisms and iterative refinement processes. Unlike autoregressive models that generate tokens sequentially from left to right, dLLMs leverage masked token prediction with full sequence context awareness, enabling them to overcome fundamental limitations such as the reversal curse \cite{berglund2024reversalcursellmstrained}. LLaDA, an 8-billion parameter model trained from scratch, rivals LLaMA3 8B performance. The masked diffusion process inherently supports in-context learning and instruction following through its non-autoregressive architecture, opening unique opportunities for embedding structured reasoning directly within the sequence, rather than relying solely on prefix prompting in autoregressive models.

\noindent \textbf{Chain-of-Thought Reasoning.}
The introduction of Chain-of-Thought (CoT) prompting \cite{wei2022chainofthought, nakano2021webgpt} has substantially enhanced reasoning accuracy in LLMs by facilitating systematic step-by-step problem decomposition. However, traditional CoT approaches are inherently constrained by their reliance on autoregressive generation and prompt-level guidance, which limits their effective integration with diffusion-based architectures. Recent advances predominantly target autoregressive models \cite{kojima2022large, gao2023synthesizing}, thereby failing to leverage the distinctive bidirectional context modeling capabilities inherent in dLLMs. To address this gap, we introduce in-place CoT prompting that seamlessly embeds reasoning steps directly into the sequence during iterative refinement, enabling fine-grained control throughout the generation.

\noindent \textbf{Efficient Inference for dLLMs.}
Diffusion LLMs inherently suffer from high computational overhead due to their iterative generation process. Current acceleration approaches for dLLMs predominantly focus on computational optimization strategies at the algorithmic level~\cite{wu2025fastdllmtrainingfreeaccelerationdiffusion, ma2025dkv, hu2025accelerating, luxembourg2025plan}. dLLM-Cache \cite{liu2025dllmcacheacceleratingdiffusionlarge} addresses this challenge through a training-free adaptive caching framework that strategically reuses stable prompt computations and employs similarity-guided partial response updates. SlowFast Sampling \cite{wei2025acceleratingdiffusionlargelanguage} further enhances inference speed by dynamically alternating between an exploratory `Slow' phase for uncertainty resolution and an aggressive `Fast' phase for confident generation, guided by certainty, convergence, and positional principles. In contrast to these algorithmic approaches, our method introduces a content-aware acceleration framework that leverages the unique bidirectional nature of dLLMs for confidence-based early exit.

\section{Methodology}\label{sec:method}

\subsection{Preliminaries}
\noindent \textbf{Masked Diffusion Large Language Models.} Masked Diffusion Large Language Models (dLLMs) implement a forward process that incrementally corrupts an input sequence $x_0$ by introducing a special \texttt{[MASK]} token. This process is controlled by a continuous time parameter $t \in [0,1]$. At each timestep $t$, the resulting sequence $x_t$ is partially masked, with each token independently replaced by \texttt{[MASK]} with probability $1 - \alpha_t$ or retained with probability $\alpha_t$. The noise schedule $\alpha_t$, which decreases monotonically with $t$, determines the corruption rate. At $t = 1$, the sequence $x_1$ is entirely masked, consisting solely of \texttt{[MASK]} tokens.

Training a masked dLLM involves a bidirectional predictor $f_\theta$ that reconstructs the original sequence $x_0$ from its corrupted counterpart $x_t$. During each training step, a timestep $t \in [0,1)$ is uniformly sampled, and tokens are masked based on the forward process defined by $\alpha_t$. The objective is to minimize the negative evidence lower bound (NELBO), an upper bound on the negative log-likelihood of the data. For masked dLLMs, the NELBO reduces to a weighted log-likelihood loss, with weights derived from $\alpha_t$~\citep{sahoo2024simple}. The popular LLaDA models use a linear noise schedule $\alpha_t = 1 - t$, in which the resulting NELBO is given by:
\begin{equation}
- \mathbb{E} \left[ \frac{1}{t} \sum_{k=1}^{|x_t|} \mathbb{I}[x_t^k = \texttt{[MASK]}] \log f_{\theta}(x_0^k \mid x_t) \right],
\label{eq:llada_loss}
\end{equation}
where $t \sim \mathcal{U}[0,1)$, $x_0 \sim p_{\text{data}}$, $x_t \sim q_{t|0}(x_t | x_0)$, $|x_t|$ represents the sequence length, $x_t^k$ denotes the $k$-th token of $x_t$, and $\mathbb{I}[\cdot]$ is an indicator function that restricts the loss to masked tokens. In contrast to BERT~\citep{devlin2018bert}, which relies on a static masking ratio and single-step token prediction, masked dLLMs employ dynamic masking probabilities and support iterative decoding from a fully masked state, enabling generative modeling.

\noindent \textbf{Inference Process of Masked dLLMs.} The inference procedure reverses the forward corruption through iterative refinement, progressively transforming a fully masked sequence into coherent output. Given an inference sequence $\mathbf{y} = (\mathbf{y}_{\text{prompt}}, \mathbf{y}_{\text{gen}})$, where $\mathbf{y}_{\text{prompt}}$ represents the initial prompt and $\mathbf{y}_{\text{gen}}$ denotes the sequence to be generated, the model maintains an intermediate state $\mathbf{y}^{(k)} = (\mathbf{y}_{\text{prompt}}, \mathbf{y}_{\text{gen}}^{(k)}) \in \mathcal{T}^L$ across $N$ discrete refinement steps, progressing from $k = N$ to $k = 0$. Here, $\mathcal{T}$ denotes the token vocabulary and $L$ represents the total sequence length. The process initializes with a fully masked generation sequence:
\begin{equation}
\mathbf{y}^{(N)} = (\mathbf{y}_{\text{prompt}}, \underbrace{\texttt{[MASK]}, \dots, \texttt{[MASK]}}_{L_{\text{gen}} \text{ times}})
\end{equation}
At each step $k \in \{N, N-1, \dots, 1\}$, the bidirectional predictor $f_\theta$ estimates the original sequence $\mathbf{y}_0$ from the current noisy state $\mathbf{y}^{(k)}$:
\begin{equation}
f_\theta(\mathbf{y}_0 \mid \mathbf{y}^{(k)})
\end{equation}
The model obtains an estimate of the clean sequence at step $k$, denoted $\hat{\mathbf{y}}_0^{(k)}$, through greedy decoding:
\begin{equation}
\hat{\mathbf{y}}_{0,i}^{(k)} = \argmax_{v \in \mathcal{T}} f_\theta(\mathbf{y}_{0,i} = v \mid \mathbf{y}^{(k)}) \quad \forall i \in \{1, \dots, L\}
\label{eq:greedy_decode_step}
\end{equation}
Subsequently, a transition function $S$ generates the next state $\mathbf{y}^{(k-1)}$ by selectively updating tokens in $\mathbf{y}^{(k)}$ based on the current estimate $\hat{\mathbf{y}}_0^{(k)}$:
\begin{equation}
\mathbf{y}^{(k-1)} = S(\hat{\mathbf{y}}_0^{(k)}, \mathbf{y}^{(k)}, k)
\label{eq:dllm_transition_step}
\end{equation}
The implementation of $S$ typically employs strategies such as confidence-based unmasking or stochastic unmasking. 

\subsection{In-Place Chain-of-Thought Prompting} 
The iterative, non-autoregressive generation paradigm of dLLMs, coupled with their inherent bidirectional attention mechanisms, enables a fundamental departure from the conventional prefix-only prompting strategies employed in autoregressive models. While autoregressive models are constrained by sequential, left-to-right generation processes, dLLMs possess the capability to consider entire sequence contexts simultaneously and enable \textit{concurrent answer accessibility}. This architectural advantage unlocks novel prompting paradigms as it allows the model to simultaneously refine reasoning steps while maintaining awareness of answer regions throughout the generation process.

Our approach leverages this distinctive capability by structuring the generation sequence $\mathbf{y}_{\text{gen}}$ into two semantically distinct sections: a \textit{thinking} section $\mathbf{y}_{\text{thinking}}$ and an \textit{answer} section $\mathbf{y}_{\text{answer}}$. This structural division is uniquely enabled by dLLMs' bidirectional nature: Unlike autoregressive models where reasoning must be sequentially generated before any answer content becomes available, dLLMs can simultaneously consider both reasoning and answer contexts during iterative refinement. Formally, we define the generation sequence as:
\begin{equation}
\mathbf{y}_{\text{gen}} = (\mathbf{y}_{\text{thinking}}, \mathbf{y}_{\text{answer}})
\end{equation}
Consequently, the complete input sequence is structured as:
\begin{equation}
\mathbf{y} = (\mathbf{y}_{\text{prompt}}, \mathbf{y}_{\text{thinking}}, \mathbf{y}_{\text{answer}})
\end{equation}

To establish clear demarcation between these functional sections, we introduce a task-specific \textit{answer indicator} positioned between the thinking and answer sections. This indicator serves as an explicit signal for the model to transition from reasoning elaboration to final response formulation.

Building upon this foundation, we further guide the model toward systematic reasoning decomposition by partitioning the thinking section $\mathbf{y}_{\text{thinking}}$ into multiple explicit reasoning steps $T=(T_1, T_2, \dots, T_{N_t})$. This is achieved by embedding explicit step-by-step reasoning templates among the masked tokens. The thinking section is initialized as:
\begin{equation}
\mathbf{y}_{\text{thinking}}^{(N)} = (\underbrace{T_1, T_2, \dots, T_{N_t}}_{N_t ~\text{reasoning steps}})
\end{equation}

This methodology of embedding structural reasoning cues directly within the generation space effectively guides the dLLM to produce explicit, traceable reasoning sequences prior to answer formulation, thereby enhancing both transparency and accuracy of the reasoning process. Crucially, this in-place approach leverages the concurrent answer accessibility of dLLMs, enabling the model to maintain holistic awareness of both reasoning development and answer formation throughout the iterative refinement process.

\subsection{Two-Phase Decoding with Early Exit Mechanism}

While our in-place chain-of-thought prompting significantly enhances reasoning capabilities, the iterative nature of dLLM inference introduces substantial computational overhead. To address this challenge, we leverage the concurrent answer accessibility of dLLMs for confidence-aware optimization. We observe a critical pattern in the confidence dynamics during iterative refinement: the model's confidence in answer tokens exhibits rapid convergence to high levels and maintains remarkable stability throughout subsequent iterations, while the thinking section continues undergoing refinement (Figure~\ref{fig:confidence_trends}). This asymmetric convergence pattern suggests that continuing refinement beyond early confidence stabilization yields diminishing returns in answer quality while incurring unnecessary computational costs.

This phenomenon reveals that models often internally converge on correct answers substantially earlier than the completion of explicit reasoning traces. Continuing iterative refinement beyond this confidence convergence point yields diminishing returns in answer quality while incurring unnecessary computational costs, as subsequent iterations primarily serve to elaborate and refine reasoning steps rather than improve final answer accuracy.

Leveraging this insight, we introduce an efficient inference strategy comprising a two-phase decoding process with a confidence-based early-exit mechanism. The core innovation lies in using a confidence threshold $\tau$ to dynamically control the transition between phases, enabling adaptive computation based on the model's internal state.

\noindent \textbf{Confidence-Based Phase Switching.} Throughout decoding, we continuously monitor the average confidence of the masked answer tokens. We first compute the confidence score for each individual answer token:
\begin{equation}
\text{confidence}_i^{(k)} = \max_{v \in \mathcal{T}} f_\theta(\mathbf{y}_{0,i} = v \mid \mathbf{y}^{(k)})
\end{equation}
Then, we calculate the average confidence across all masked answer tokens:
\begin{equation}
\text{avg\_confidence}_{\text{answer}}^{(k)} = \frac{1}{L_{\text{answer}}} \sum_{i \in \text{answer indices}} \text{confidence}_i^{(k)}
\end{equation}
This confidence threshold $\tau$ serves as the decisive criterion for phase transition: when $\text{avg\_confidence}_{\text{answer}}^{(k)} \geq \tau$, we trigger an early exit from the reasoning phase and immediately transition to answer generation.

\noindent \textbf{Phase 1: Reasoning Phase.} During this initial phase, we focus on generating the thinking trace ($\mathbf{y}_{\text{thinking}}$) while maintaining all answer tokens in their masked state. The transition function $S$ (Equation~\ref{eq:dllm_transition_step}) unmasks tokens solely within $\mathbf{y}_{\text{thinking}}$, guided by model confidence scores. Crucially, we simultaneously monitor the answer confidence for early exit detection. When the confidence threshold is reached, this phase terminates immediately.

\noindent \textbf{Phase 2: Answer Generation Phase.} This phase is triggered exclusively when $\text{avg\_confidence}_{\text{answer}}^{(k)} \geq \tau$. Upon transition, we perform a single-step decoding operation to reveal the complete answer sequence $\mathbf{y}_{\text{answer}}$. This dynamic phase switching eliminates redundant computation while preserving accuracy, as the model has demonstrated enough confidence in the final answer. A detailed algorithm overview of the ICE framework is provided in the appendix.

\section{Experiments}\label{sec:exp}

\subsection{Experimental Setup}

We evaluate ICE on diverse benchmarks: GSM8K~\cite{cobbe2021traininggsm8k} and MATH~\cite{hendrycks2021measuringmath} for mathematical reasoning, and MMLU~\cite{hendrycks2020measuringmmlu} and GPQA~\cite{rein2023gpqa} for knowledge-intensive reasoning. We conduct comprehensive evaluations using two representative dLLMs: LLaDA-8B-Instruct~\cite{nie2025large} and LLaDA-1.5~\cite{Zhu2025LLaDA1V}, with the Language Model Evaluation Harness framework~\cite{eval-harness} on 8 NVIDIA H100 GPUs. We compare ICE against two baseline approaches: (1) \emph{Vanilla}: the standard autoregressive generation approach, and (2) \emph{Prefix CoT}: traditional chain-of-thought prompting as input prefixes. ICE offers two operational modes: ICE-SP (speed-prioritized) for maximum acceleration and ICE-PP (performance-prioritized) for superior accuracy, which are configured by the hyperparameters.

\begin{table*}[t!]
  \centering
  \begin{tabular}{c|l|llc|llc}
     \toprule
    \multirow{2}{*}{\bf Task} & \multirow{2}{*}{\bf Method} & \multicolumn{3}{c|}{\bf LLaDA-8B-Instruct} & \multicolumn{3}{c}{\bf LLaDA-1.5} \\
    \cmidrule(lr){3-5} \cmidrule(lr){6-8}
    & & {\bf Accuracy~$\uparrow$} & {\bf Latency (s)~$\downarrow$} & {\bf Speedup~$\uparrow$} & {\bf Accuracy~$\uparrow$} & {\bf Latency (s)~$\downarrow$} & {\bf Speedup~$\uparrow$}\\
  \midrule
    \multirow{4}{*}{GSM8K} & Vanilla & 46.55 & 1461.83 & 1.00$\times$ & 45.19 & 3376.21 & 1.00$\times$ \\
                           & ~+~Prefix CoT & 40.49$_{-6.06}$ & 1443.02$_{-18.81}$ & 1.01$\times$ & 34.95$_{-10.24}$ & 3555.10$_{+178.89}$ & 0.95$\times$ \\
                           & \cellcolor{gray!15}ICE-SP & \cellcolor{gray!15}46.01$_{-0.54}$ & \cellcolor{gray!15}\textbf{354.86}$_{-1106.97}$ & \cellcolor{gray!15}\textbf{4.12$\times$} & \cellcolor{gray!15}45.56$_{+0.37}$ & \cellcolor{gray!15}\textbf{1050.48}$_{-2325.73}$ & \cellcolor{gray!15}\textbf{3.21}$\times$\\
                           & \cellcolor{gray!15}ICE-PP & \cellcolor{gray!15}\textbf{63.84}$_{+17.29}$ & \cellcolor{gray!15}874.53$_{-587.30}$ & \cellcolor{gray!15}1.67$\times$ & \cellcolor{gray!15}\textbf{58.22}$_{+13.03}$ & \cellcolor{gray!15}2221.24$_{-1154.97}$ & \cellcolor{gray!15}1.52$\times$\\
  \midrule
    \multirow{4}{*}{MATH} & Vanilla & 23.88 & 3570.13 & 1.00$\times$ & 22.74 & 10018.06 & 1.00$\times$ \\
                          & ~+~Prefix CoT & 22.26$_{-1.62}$ & 3517.61$_{-52.52}$ & 1.01$\times$ & 19.76$_{-2.98}$ & 10034.73$_{+16.67}$ & 1.00$\times$ \\
                          & \cellcolor{gray!15}ICE-SP & \cellcolor{gray!15}23.68$_{-0.20}$ & \cellcolor{gray!15}\textbf{2132.79}$_{-1437.34}$ & \cellcolor{gray!15}\textbf{1.67$\times$} & \cellcolor{gray!15}22.74$_{0.00}$ & \cellcolor{gray!15}\textbf{5885.87}$_{-4132.19}$ & \cellcolor{gray!15}\textbf{1.70}$\times$ \\
                          & \cellcolor{gray!15}ICE-PP & \cellcolor{gray!15}\textbf{26.88}$_{+3.00}$ & \cellcolor{gray!15}3155.99$_{-414.14}$ & \cellcolor{gray!15}1.13$\times$ & \cellcolor{gray!15}\textbf{25.28}$_{+2.54}$ & \cellcolor{gray!15}8774.40$_{-1243.66}$ & \cellcolor{gray!15}1.14$\times$ \\
  \midrule
    \multirow{3}{*}{GPQA} & Vanilla & 27.46 & 970.84 & 1.00$\times$ & 28.13 & 2156.62 & 1.00$\times$ \\
                          & ~+~Prefix CoT & 27.68$_{+0.22}$ & 1043.77$_{+72.93}$ & 0.93$\times$ & 27.90$_{-0.23}$ & 2234.75$_{+78.13}$ & 0.97$\times$ \\
                          & \cellcolor{gray!15}ICE & \cellcolor{gray!15}\textbf{32.37}$_{+4.91}$ & \cellcolor{gray!15}\textbf{50.45}$_{-920.39}$ & \cellcolor{gray!15}\textbf{19.24}$\times$ & \cellcolor{gray!15}\textbf{33.70}$_{+5.57}$ & \cellcolor{gray!15}\textbf{51.24}$_{-2105.38}$ & \cellcolor{gray!15}\textbf{42.08}$\times$ \\

  \midrule
    \multirow{3}{*}{MMLU} & Vanilla & 49.67 & 19396.79 & 1.00$\times$ & 60.35 & 47795.97 & 1.00$\times$ \\
                          & ~+~Prefix CoT & 51.22$_{+1.55}$ & 19469.64$_{+72.85}$ & 1.00$\times$ & 60.11$_{-0.24}$ & 48298.11$_{+502.14}$ & 0.99$\times$ \\
                          & \cellcolor{gray!15}ICE & \cellcolor{gray!15}\textbf{62.77}$_{+13.10}$ & \cellcolor{gray!15}\textbf{145.75}$_{-19251.04}$ & \cellcolor{gray!15}\textbf{133.08}$\times$ & \cellcolor{gray!15}\textbf{61.09}$_{+0.74}$ & \cellcolor{gray!15}\textbf{172.79}$_{-47623.18}$ & \cellcolor{gray!15}\textbf{276.67}$\times$\\
  \bottomrule
  \end{tabular}
  \caption{
    Performance comparison of ICE on different tasks. Experiments are conducted using LLaDA-8B-Instruct 
    (length 256) and LLaDA-1.5 (length 512), with block length and generation steps matching generation length. \emph{Prefix CoT} refers to the vanilla with CoT prompting as prefixes. ICE operates in SP mode for inference acceleration and PP mode for enhanced accuracy. For MMLU and GPQA, a unified configuration achieves both superior accuracy improvements and efficiency gains.
  }
  \label{tab:main_results}
\end{table*}

\subsection{Main Results}
Our main results are summarized in Table~\ref{tab:main_results}.

\noindent \textbf{Mathematical Reasoning Tasks.}
For complex mathematical reasoning problems, ICE-PP achieves substantial accuracy improvements: +17.29\% and +13.03\% improvements on GSM8K with 1.67$\times$ and 1.52$\times$ speedups, and +3.00\% and +2.54\% improvements on MATH while maintaining computational efficiency. ICE-SP delivers exceptional efficiency (4.12$\times$ and 3.21$\times$ on GSM8K, 1.67$\times$ and 1.70$\times$ on MATH) with near-zero accuracy loss, demonstrating versatile optimization capabilities.

\noindent \textbf{Knowledge-Intensive Tasks.}
For knowledge-intensive reasoning tasks, ICE achieves simultaneous accuracy improvements and remarkable efficiency gains. GPQA shows substantial improvements of +4.91\% and +5.57\% with extraordinary speedups of 19.24$\times$ and 42.08$\times$, demonstrating effectiveness on tasks requiring deep domain expertise. MMLU achieves +13.10\% and +0.74\% improvements with exceptional speedups up to 133.08$\times$ and 276.67$\times$, with the larger accuracy gain for LLaDA Instruct indicating particular benefits for non-preference-optimized models. These results suggest that diverse reasoning queries can be effectively resolved through early-layer representations without requiring explicit speed-accuracy trade-offs.

\begin{figure}[t]
    \centering
    \includegraphics[width=0.9\columnwidth]{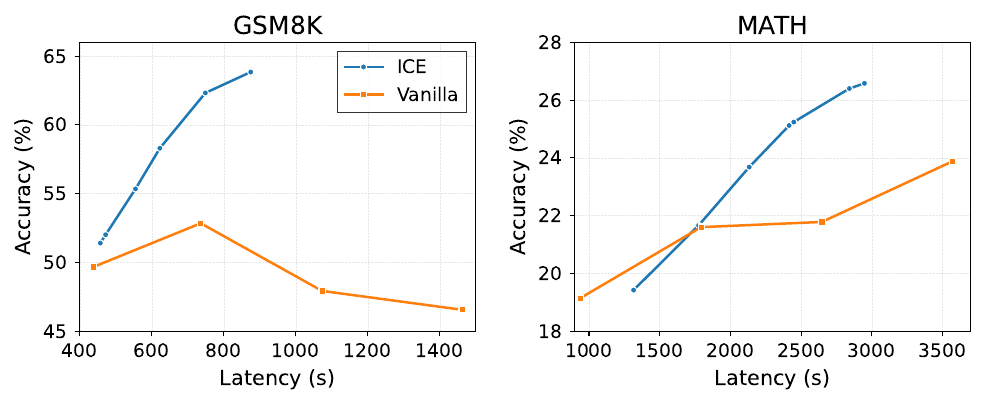} 
    \caption{Latency-accuracy trade-off comparison between ICE and vanilla baselines. ICE demonstrates superior Pareto frontiers on both GSM8K and MATH datasets.}
    \label{fig:latency_accuracy}
\end{figure}

\noindent \textbf{Latency-Accuracy Trade-off Comparison.} Figure~\ref{fig:latency_accuracy} compares ICE with vanilla baselines that achieve different trade-off points by fixing output length and adjusting generation steps. ICE establishes superior Pareto frontiers on both datasets: reducing latency by 2-4$\times$ on GSM8K while maintaining accuracy, and achieving both lower latency and higher accuracy on MATH.

\begin{table}[t]
    \centering
    % \small
    \begin{tabular}{c|lccc}
    \toprule
    \textbf{Task} & \textbf{Method} & \textbf{Acc.} & \textbf{Latency} & \textbf{Speedup} \\
    \midrule
    \multirow{5}{*}{GSM8K} & Vanilla & 46.55 & 1461.83 & 1.00$\times$ \\
                          & ICE-SP & 46.01 & 354.86 & 4.12$\times$ \\
                          & \cellcolor{gray!15}~+~Cache & \cellcolor{gray!15}\textbf{46.17} & \cellcolor{gray!15}\textbf{239.83} & \cellcolor{gray!15}\textbf{6.10}$\times$ \\
                          & ICE-PP & \textbf{63.84} & 874.53 & 1.67$\times$ \\
                          & \cellcolor{gray!15}~+~Cache & \cellcolor{gray!15}61.56 & {\cellcolor{gray!15}\textbf{592.73}} & {\cellcolor{gray!15}\textbf{2.47}$\times$} \\
    \midrule
    \multirow{5}{*}{MATH} & Vanilla & 23.88 & 3570.13 & 1.00$\times$ \\
                         & ICE-SP & 23.68 & 2132.79 & 1.67$\times$ \\
                         & \cellcolor{gray!15}~+~Cache & \cellcolor{gray!15}\textbf{23.72} & \cellcolor{gray!15}\textbf{1771.15} & \cellcolor{gray!15}\textbf{2.02}$\times$ \\
                         & ICE-PP & \textbf{26.88} & 3155.99 & 1.13$\times$ \\
                         & \cellcolor{gray!15}~+~Cache & \cellcolor{gray!15}25.94 & \cellcolor{gray!15}\textbf{2684.77} & \cellcolor{gray!15}\textbf{1.33}$\times$ \\
    \bottomrule
    \end{tabular}
    \caption{Compatibility evaluation of ICE with dLLM-Cache on LLaDA-8B-Instruct across GSM8K and MATH.}
    \label{tab:cache}
\end{table}

\noindent \textbf{Compatibility with dLLM-Cache.}
To validate ICE's compatibility with existing optimization techniques, we evaluate its integration with dLLM-Cache. Table~\ref{tab:cache} demonstrates that ICE maintains its effectiveness when combined with caching mechanisms, achieving substantial additional speedups while preserving accuracy. For ICE-SP, cache acceleration enhances speedup from 4.12$\times$ to 6.10$\times$ on GSM8K and from 1.67$\times$ to 2.02$\times$ on MATH. The ICE-PP similarly benefits from cache integration, achieving 2.47$\times$ speedup on GSM8K and 1.33$\times$ on MATH. The accuracy degradation introduced by caching is minimal (typically $<$2\%), confirming that our structured reasoning approach remains robust with complementary optimization strategies.

\subsection{Ablation Study}
\label{sec:ablation}

To gain deeper insights into ICE's design choices, we conduct ablation studies focusing on the key hyperparameters and architectural decisions that drive the performance.

\begin{table}[t]
\centering
% \small
\begin{tabular}{@{}l@{\hspace{0.2cm}}c@{\hspace{0.2cm}}c@{\hspace{0.2cm}}c@{\hspace{0.2cm}}c@{}}
\toprule
\textbf{Task} & \textbf{Vanilla} & \textbf{+ Segment} & \textbf{+ Structured} & \textbf{+ Early Exit} \\
 & & \textbf{(T/A)} & \textbf{Thinking} & \textbf{(ICE)} \\
\midrule
\rowcolor{gray!20}
\multicolumn{5}{l}{\textbf{\textit{LLaDA-8B-Instruct}}} \\
GSM8K & 46.55 & 55.95 & \textbf{64.37} & 63.84 \\
MATH & 23.88 & 24.18 & \textbf{26.88} & \textbf{26.88} \\
GPQA & 27.68 & 29.91 & 31.70 & \textbf{32.37} \\
\midrule
\rowcolor{gray!20}
\multicolumn{5}{l}{\textbf{\textit{LLaDA-1.5}}} \\
GSM8K & 45.19 & 52.54 & 55.04 & \textbf{58.22} \\
MATH & 22.74 & 23.74 & 24.64 & \textbf{25.28} \\
GPQA & 27.90 & 29.91 & 30.13 & \textbf{33.70} \\
\bottomrule
\end{tabular}
\caption{Ablation study on the core components of ICE.}
\label{tab:ablation_study}
\end{table}

\noindent \textbf{Impact of Core Components.} Table~\ref{tab:ablation_study} demonstrates the incremental contributions of ICE's architectural components. The initial thinking/answer segmentation mechanism (+ Segment) provides consistent improvements across all benchmarks, yielding +9.40\% and +7.35\% gains on GSM8K for LLaDA-8B-Instruct and LLaDA-1.5 respectively, validating the fundamental importance of explicit reasoning structure in discrete masked language models. The introduction of structured thinking subdivision further enhances performance, contributing an additional +8.42\% and +2.50\% on GSM8K, demonstrating that fine-grained decomposition of reasoning processes enables more effective utilization of the masked generation paradigm. The confidence-based early exit mechanism (ICE) exhibits task-dependent behavior: while showing marginal accuracy trade-offs on GSM8K (-0.53\% for LLaDA-8B-Instruct), it provides substantial improvements on knowledge-intensive tasks like GPQA (+0.67\% and +3.57\%), indicating that the early exit strategy is particularly effective when reasoning complexity varies significantly across tasks.

% \begin{figure}[t!]
%     \centering
%     \includegraphics[width=\columnwidth]{figures/steps_vertical.pdf} 
%     \caption{Ablation study on the number of explicit thoughts ($N_t$) across GSM8K, MATH and GPQA benchmarks.}
%     \label{fig:thought_numbers}
% \end{figure}
\begin{figure}[t]
    \centering
    \includegraphics[width=0.9\columnwidth]{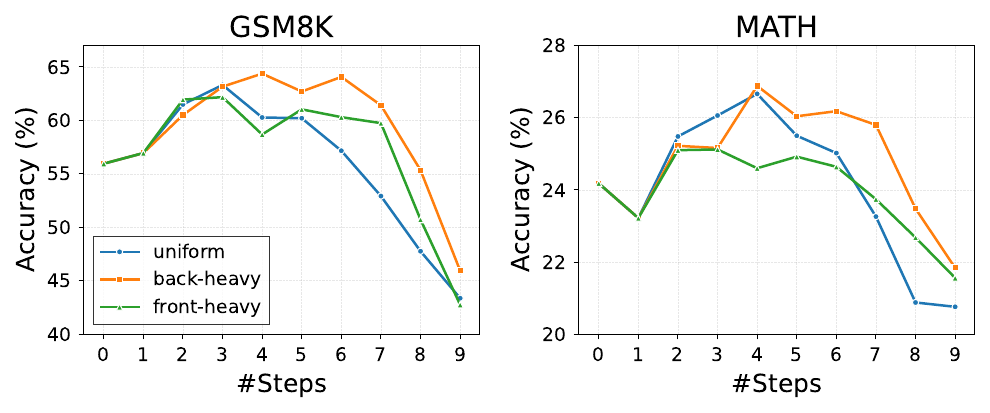} 
    \caption{Ablation study on reasoning steps ($N_t$).}
    \label{fig:thought_numbers}
\end{figure}

\noindent \textbf{Effect of Reasoning Steps ($N_t$).} Figure~\ref{fig:thought_numbers} demonstrates the critical relationship between reasoning step granularity and model performance. Our systematic analysis reveals distinct optimal operating points for different task domains: GSM8K achieves peak performance at $N_t = 3$ with approximately 58-60\% accuracy, while MATH demonstrates optimal results at $N_t = 4$ with 25-26\% accuracy. This finding suggests that mathematical reasoning tasks benefit from moderate decomposition depths that balance reasoning granularity with computational overhead. Notably, excessive subdivision ($N_t > 6$) consistently degrades performance across all benchmarks, indicating that overly fine-grained reasoning steps may introduce noise and computational inefficiency. For knowledge-intensive tasks like GPQA, the framework maintains stable performance across a broader range of thought numbers (28-31\% accuracy), demonstrating robustness to hyperparameter variations.

\noindent \textbf{Mask Token Allocation Strategies.} We evaluate three distinct strategies for distributing mask tokens across reasoning steps: \emph{uniform} allocation maintains equal token budgets for each thought, \emph{front-heavy} allocation prioritizes initial reasoning steps, and \emph{back-heavy} allocation concentrates computational resources on final reasoning phases. The experimental results demonstrate that back-heavy and front-heavy strategies consistently outperform uniform allocation, particularly in lower thought number regimes ($N_t \leq 4$). This finding suggests that strategic token concentration yields superior performance compared to equal resource distribution, highlighting the importance of adaptive computational allocation in structured reasoning frameworks.

\begin{figure}[t]
    \centering
    \includegraphics[width=0.9\columnwidth]{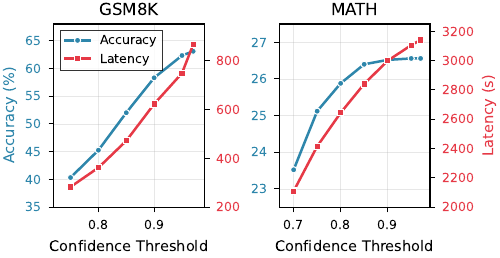} 
    \caption{Ablation study on the confidence threshold $\tau$.}
    \label{fig:confidence_threshold}
\end{figure}

%%%%%% ADD 0 Thoughts
% Remove GPQA

\noindent \textbf{Confidence Threshold Analysis.} The confidence threshold $\tau$ serves as a pivotal hyperparameter governing the speed-accuracy trade-off in our early exit mechanism. Figure~\ref{fig:confidence_threshold} systematically illustrates this relationship across mathematical reasoning benchmarks (GSM8K and MATH). Our analysis reveals a clear pattern: lower thresholds aggressively prioritize computational efficiency at the expense of accuracy, while higher thresholds emphasize accuracy preservation with reduced speedup benefits. Critically, moderate thresholds achieve the optimal balance, delivering substantial accuracy improvements while maintaining significant computational efficiency gains, thereby validating the effectiveness of our confidence-based early exit strategy. For knowledge-intensive tasks (GPQA and MMLU), we observe that excessively high confidence thresholds yield diminishing returns, with accuracy plateauing rather than improving further. This finding underscores the importance of task-adaptive threshold selection, where reasoning complexity should inform the optimal confidence calibration for maximum effectiveness.

\section{Discussion}
\label{sec:discussion}

% Our comprehensive evaluation of ICE reveals important insights into structured reasoning paradigms in diffusion language models, offering deeper understanding of prompting methodologies for non-autoregressive generation.

\subsection{In-Place Prompting: A New Paradigm for dLLMs}

Our work repositions chain-of-thought from a sequential pre-computation into a dynamic, symbiotic component of the dLLM's iterative refinement process. By embedding prompts directly within the generation canvas, we move beyond the prefix-only constraints of autoregressive models. This integration fosters a co-refinement dynamic where the reasoning trace and the final answer evolve in parallel, mutually informing one another throughout generation steps.

This paradigm shift is unique to dLLMs. Unlike autoregressive models where a generated reasoning step is immutable, our approach allows the model to revisit and refine its entire line of thought in light of the emerging answer. The reasoning templates ($T_1, \ldots, T_{N_t}$) act less like rigid instructions and more like a flexible scaffold that guides the refinement process while adapting to the model's evolving understanding of the problem. This suggests a fundamental compatibility between structured problem decomposition and the core mechanics of diffusion models, transforming reasoning from a static prerequisite into a concurrently optimized process. The implications suggest that dLLMs could be architected with internal scaffolds for other complex tasks like constrained generation and planning.

\subsection{Internal Dynamics in dLLMs}

\noindent \textbf{Decoupling Solution Finding from Explanation Generation.}
Our findings reveal that dLLMs effectively decouple solution finding from explanation generation. The rapid confidence convergence in answer sections, while reasoning traces remain unstable, indicates that models stabilize final answers before completing narrative justifications. This contrasts with AR models where solutions and explanations form sequential chains, which enables our early-exit mechanism to leverage a fundamental property of dLLM rather than merely providing computational efficiency.

\begin{figure}[t]
    \centering
    \includegraphics[width=0.9\columnwidth]{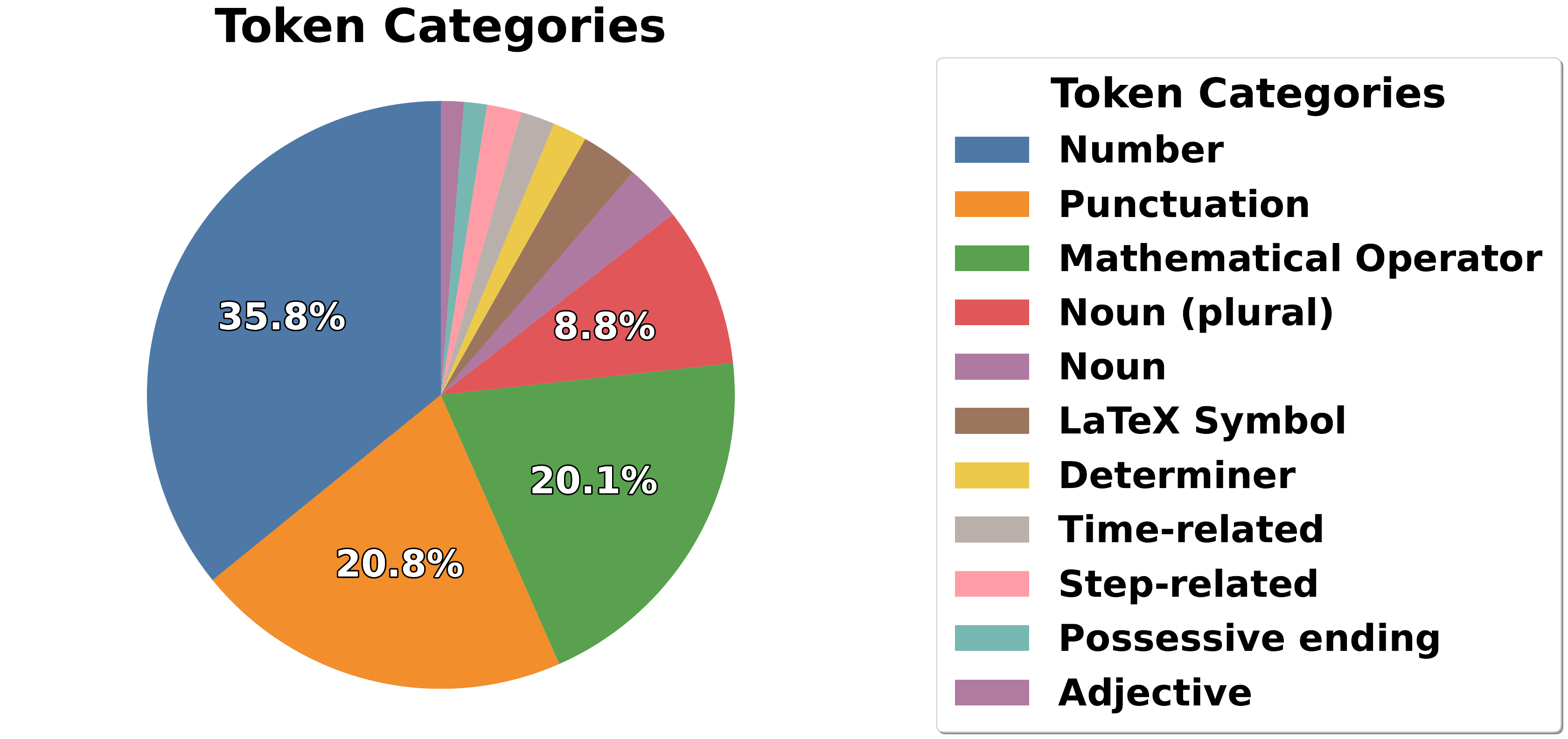} 
    \caption{Statistics of token categories decoded when confidence rapidly changes during generation on GSM8K.}
    \label{fig:confidence_jump}
\end{figure}

\noindent \textbf{Token-Level Confidence Dynamics.}
Token-level analysis reveals that confidence maturation occurs through decisive jumps rather than gradual increases. Figure~\ref{fig:confidence_jump} presents the distribution of token categories that are decoded when confidence rapidly changes during GSM8K generation. The statistics show these critical shifts are primarily driven by numerical token stabilization. The model first anchors quantitative results, then refines punctuation and mathematical operators to solidify reasoning syntax. This hierarchical convergence process locks core results before constructing explanatory scaffolding. Interestingly, this finding echoes observations from SepLLM~\cite{chen2025sepllm}, which identified high attention scores for punctuation tokens in autoregressive LLMs, suggesting that structural tokens play crucial roles across different model architectures.

\section{Conclusion}

We introduce ICE, a novel framework that enhances both reasoning capabilities and inference efficiency in dLLMs through in-place CoT prompting and two-phase decoding with confidence-aware early exit mechanism. Our approach leverages dLLMs' natural advantages of bidirectional attention, achieving significant improvements with up to 17.29\% accuracy gains and 4.12$\times$ speedup on GSM8K, and up to 276.67$\times$ acceleration on MMLU. This work demonstrates that architectural alignment between reasoning patterns and generation mechanisms can yield synergistic benefits, transforming iterative refinement from computational burden into architectural advantage and establishing a new paradigm for efficient inference in non-autoregressive language models.

\clearpage
\bibliography{aaai2026}

\clearpage
\appendix

\section*{Appendix}

\section{Implementation Details}

\noindent \textbf{Datasets.} We evaluate ICE on four diverse benchmarks spanning mathematical and knowledge-intensive tasks. For mathematical reasoning, we use \textbf{GSM8K}~\cite{cobbe2021traininggsm8k}, a dataset of 8,500 grade school math word problems requiring multi-step reasoning, and \textbf{MATH}~\cite{hendrycks2021measuringmath}, a challenging collection of high school competition mathematics problems across algebra, geometry, and other domains. For knowledge-intensive reasoning, we employ \textbf{MMLU}~\cite{hendrycks2020measuringmmlu}, a comprehensive benchmark covering 57 subjects from elementary mathematics to advanced professional topics, and \textbf{GPQA}~\cite{rein2023gpqa}, a graduate-level Google-proof Q\&A dataset designed to evaluate advanced reasoning in biology, physics, and chemistry. These benchmarks collectively assess both the breadth and depth of reasoning capabilities across different domains and difficulty levels.

\noindent \textbf{Models and Evaluation Protocol.} We conduct comprehensive evaluations using two representative dLLMs: LLaDA-8B-Instruct, an 8B-parameter masked discrete 
diffusion model trained from scratch, and LLaDA-1.5, which incorporates Variance-Reduced Preference Optimization (VRPO) for improved human preference alignment. To ensure consistent and reproducible results, we employ the widely adopted Language Model Evaluation Harness framework~\cite{eval-harness} across all benchmarks.

\section{Detailed Algorithm}
The detailed pseudo code of ICE is shown in Algorithm~\ref{alg:ICE}.

\begin{algorithm}[h]
    \caption{ICE: In-Place CoT Prompting with Early Exit}
    \label{alg:ICE}
    \begin{algorithmic}[1]
    \REQUIRE Prompt $\mathbf{y}_{\text{prompt}}$, thinking template $T = \{T_1, T_2, \ldots, T_{N_t}\}$, answer indicator, confidence threshold $\tau$, max steps $N$
    \ENSURE Generated sequence $\hat{\mathbf{y}}_{\text{final}}$
    
    \STATE \textbf{// Phase 1: Initialize structured sequence}
    \STATE $\mathbf{y}_{\text{thinking}}^{(N)} \leftarrow (T_1, T_2, \ldots, T_{N_t})$ \COMMENT{Insert thinking templates}
    \STATE $\mathbf{y}_{\text{answer}}^{(N)} \leftarrow (\texttt{[MASK]}, \ldots, \texttt{[MASK]})$ \COMMENT{Mask all answer tokens}
    \STATE $\mathbf{y}^{(N)} \leftarrow (\mathbf{y}_{\text{prompt}}, \mathbf{y}_{\text{thinking}}^{(N)}, \mathbf{y}_{\text{answer}}^{(N)})$
    \STATE $k \leftarrow N$, $\text{phase} \leftarrow \text{reasoning}$
    
    \WHILE{$k > 0$ \AND $\text{phase} = \text{reasoning}$}
        \STATE \textbf{// Generate prediction for current step}
        \STATE $\hat{\mathbf{y}}_0^{(k)} \leftarrow \argmax_{v} f_\theta(\mathbf{y}_{0} = v \mid \mathbf{y}^{(k)})$ \COMMENT{Eq.~\ref{eq:greedy_decode_step}}
        
        \STATE \textbf{// Compute confidence scores}
        \FOR{$i \in \{1, \ldots, L\}$}
            \STATE $\text{confidence}_i^{(k)} \leftarrow \max_{v \in \mathcal{T}} f_\theta(\mathbf{y}_{0,i} = v \mid \mathbf{y}^{(k)})$ \COMMENT{Eq. 5}
        \ENDFOR
        
        \STATE \textbf{// Check early exit condition}
        \STATE $\text{avg\_conf}_{\text{answer}}^{(k)} \leftarrow \frac{1}{L_{\text{answer}}} \sum_{i \in \text{answer indices}} \text{confidence}_i^{(k)}$ \COMMENT{Eq. 7}
        
        \IF{$\text{avg\_conf}_{\text{answer}}^{(k)} > \tau$}
            \STATE $\text{phase} \leftarrow \text{answer\_generation}$ \COMMENT{Early exit triggered}
            \STATE \textbf{break}
        \ENDIF
        
        \STATE \textbf{// Update only thinking section}
        \STATE $\mathbf{y}^{(k-1)} \leftarrow S_{\text{thinking}}(\hat{\mathbf{y}}_0^{(k)}, \mathbf{y}^{(k)}, k)$ \COMMENT{Selective unmasking}
        \STATE $k \leftarrow k - 1$
    \ENDWHILE
    
    \STATE \textbf{// Phase 2: Answer generation}
    \IF{$\text{phase} = \text{reasoning}$}
        \STATE $\text{phase} \leftarrow \text{answer\_generation}$ \COMMENT{Normal termination}
    \ENDIF
    
    \STATE \textbf{// Single-step answer decoding}
    \STATE $\hat{\mathbf{y}}_{\text{final}} \leftarrow \argmax_{v} f_\theta(\mathbf{y}_{\text{answer}} = v \mid \mathbf{y}_{\text{current}})$
    \STATE \textbf{return} $\hat{\mathbf{y}}_{\text{final}}$
    
    \end{algorithmic}
\end{algorithm}

\end{document}